\documentclass[sigconf]{acmart}
\usepackage{caption}
\usepackage{subcaption}
\usepackage{graphicx}
\usepackage{float}
\usepackage{booktabs}

\AtBeginDocument{%
  }

\setcopyright{acmlicensed}
\copyrightyear{2025}
\acmYear{2025}
\acmDOI{10.1145/3714394.3757253}
\acmConference[UbiComp Companion '25] {Companion of the the 2025 ACM International Joint Conference on Pervasive and Ubiquitous Computing}{October 12--16, 2025}{Espoo, Finland.}
\acmBooktitle{Companion of the the 2025 ACM International Joint Conference on Pervasive and Ubiquitous Computing (UbiComp Companion '25), October 12--16, 2025, Espoo, Finland}
\acmISBN{979-8-4007-1477-1/25/10}

\begin{document}

\title{Feasibility of In-Ear Single-Channel ExG for Wearable Sleep~Monitoring in Real-World Settings}

\author{Philipp Lepold}
\authornote{Both authors contributed equally to this research.}
\email{philipp.lepold@kit.edu}
\affiliation{%
  \institution{Karlsruhe Institute of Technology}
  \city{Karlsruhe}
  \country{Germany}
}

\author{Jonas Leichtle}
\authornotemark[1]
\email{leichtle@teco.edu}
\affiliation{%
  \institution{Karlsruhe Institute of Technology}
  \city{Karlsruhe}
  \country{Germany}
}

\author{Tobias Röddiger}
\email{tobias.roeddiger@kit.edu}
\affiliation{%
  \institution{Karlsruhe Institute of Technology}
  \city{Karlsruhe}
  \country{Germany}
}

\author{Michael Beigl}
\email{michael.beigl@kit.edu}
\affiliation{%
  \institution{Karlsruhe Institute of Technology}
  \city{Karlsruhe}
  \country{Germany}
}

\renewcommand{\shortauthors}{Philipp Lepold, Jonas Leichtle, Tobias Röddiger, and Michael Beigl}

\begin{abstract}
Automatic sleep staging typically relies on gold-standard EEG setups, which are accurate but obtrusive and impractical for everyday use outside sleep laboratories. This limits applicability in real-world settings, such as home environments, where continuous, long-term monitoring is needed. Detecting sleep onset is particularly relevant, enabling consumer applications (e.g. automatically pausing media playback when the user falls asleep).
Recent research has shown correlations between in-ear EEG and full-scalp EEG for various phenomena, suggesting wearable, in-ear devices could allow unobtrusive sleep monitoring. We investigated the feasibility of using single-channel in-ear electrophysiological (ExG) signals for automatic sleep staging in a wearable device by conducting a sleep study with 11~participants (mean age: 24), using a custom earpiece with a dry eartip electrode (Dätwyler SoftPulse) as a measurement electrode in one ear and a reference in the other. Ground truth sleep stages were obtained from an Apple Watch Ultra, validated for sleep staging. Our system achieved 90.5\% accuracy for binary sleep detection (\textit{Awake} vs. \textit{Asleep}) and 65.1\% accuracy for four-class staging (\textit{Awake}, \textit{REM}, \textit{Core}, \textit{Deep}) using leave-one-subject-out validation. These findings demonstrate the potential of in-ear electrodes as a low-effort, comfortable approach to sleep monitoring, with applications such as stopping podcasts when users fall asleep.


\end{abstract}



\maketitle

\section{Introduction}
Sleep and its quality play a pivotal role in human well-being as we spend one-third of our lives sleeping \cite{chaput_sleeping_2018}. 
It is essential for various critical functions of the human body, including neural development, learning, memory, and the removal of toxins \cite{mukherjee_official_2015}. Poor sleep and sleep-related disorders can negatively impact health and lead to adverse outcomes. Insomnia, the most common sleep disorder, was ranked 11th on the list of Global Burden of Mental, Neurological, and Substance-Use (MNS) Disorders \cite{collins_grand_2011}. Unfortunately, many people with sleep disorders are undiagnosed and remain untreated \cite{mukherjee_official_2015}.
Therefore, lowering the barrier of accurate sleep measurements is of great significance both in diagnostic medicine and in sleep research. The gold standard of sleep measurements today is polysomnography (PSG), which is performed in a sleep clinic and evaluated by trained professionals \cite{rundo_polysomnography_2019}. During PSG, various parameters are measured during sleep, including brain activity (EEG), eye movements (EOG), muscle activity (EMG), airflow, and many others. While highly accurate, PSG is impractical for everyday use due to its complexity, cost, and the need for professional supervision. This limits its accessibility, especially in long-term or home-based monitoring.\\
In-ear EEG devices have emerged as a promising alternative, offering a more practical and user-friendly way to monitor brain activity. While prior research has largely focused on the technical feasibility of such devices and signal quality in controlled settings, there is limited understanding of how in-ear EEG sleep monitoring performs in real-world environments with non-expert users.\\
In this work, we show that sleep onset can be detected only with in-ear electrodes, in realistic settings, achieving promising performance while remaining accessible and user-friendly. By focusing on a practical use case and leveraging an open-source, wearable platform, our work takes a step towards closing the gap between research-grade in-ear EEG and consumer-ready, in-home sleep monitoring applications.

\section{Related Work}

\subsection{Wearables in Sleep Tracking}

Wearable devices have become ubiquitous in consumer health tracking and are increasingly used in sleep research. Commercial devices such as the Apple Watch\footnote{\url{https://www.apple.com/watch/}}or the Oura Ring\footnote{\url{https://ouraring.com/de/product/rings/oura-gen3}} offer basic sleep tracking capabilities based primarily on actigraphy and photoplethysmography (PPG). While these systems provide some insight into sleep-wake cycles, their performance in staging sleep, especially distinguishing between REM and deep sleep, is limited \cite{de_zambotti_wearable_2019}.

More advanced systems, such as clinically validated actigraphs, like the Ametris ActiGraph\footnote{\url{https://theactigraph.com/sleep}}, and EEG-based headbands, such as the Beacon Signals Waveband\footnote{\url{https://beacon.bio/dreem-headband/}}, provide improved accuracy and are used in research or clinical contexts. However, they are often expensive, obtrusive, or restricted to laboratory environments. There remains a need for wearable systems that can achieve reliable sleep staging in real-world, at-home conditions using accessible and unobtrusive hardware.


\subsection{In-Ear EEG}
Ear-centered EEG (ear-EEG) systems have emerged as a promising alternative to conventional scalp EEG, offering a less obtrusive form factor and improved comfort for long-term use. \citet{looney_--ear_2012} introduced one of the first prototypes for in-ear EEG, demonstrating that signals recorded from earpieces could approximate scalp EEG. Their work validated the concept but relied on wet electrodes, which are unsuitable for extended, unsupervised use.

Subsequent advances focused on improving wearability and usability. \citet{kappel_dry-contact_2019} introduced dry-contact electrodes in custom-fitted earpieces, enabling more practical use. \citet{mikkelsen_accurate_2019} demonstrated accurate sleep staging using machine learning applied to in-ear EEG signals collected over 80 nights. Their system achieved a Cohen's kappa of 0.73 compared to manually scored PSG. However, this system also relied on individualized earpieces and laboratory-grade PSG for validation.

\citet{tabar_at-home_2023} extended this work by evaluating a generic, non-customized ear-EEG device. Their findings suggest that ear-EEG can be used more broadly, but the study remained confined to short-term recordings and controlled environments. Similarly, \citet{nakamura_hearables_2020} and \citet{nguyen_lightweight_2016} developed in-ear sensors using flexible or foam-based electrodes, validated against scalp EEG and full PSG. These systems employed classical machine learning models (support vector machines and random forests) and achieved substantial agreement with expert annotations (Cohen's kappa up to 0.61, and classification accuracy up to 95\%).

Despite these advances, most ear-EEG studies share several limitations: they rely on proprietary or custom-molded hardware, require expert-guided installation, and are typically evaluated in laboratory or semi-controlled environments. As a result, their applicability in fully unsupervised, at-home settings remains limited. Moreover, many studies focus solely on EEG signals, whereas combining EEG with other modalities, such as heart rate variability and actigraphy, could enhance performance in real-world conditions. Open-source, reproducible systems that can operate with affordable, off-the-shelf hardware remain scarce in this space.

\section{Methodology}

\subsection{Open-Source ExG Device}
We're using the open-source platform OpenEarable ExG \cite{lepold_openearable_2024} for measuring biopotentials, due to it's simple setup, wearable design and wireless connectivity, which is not the case for virtually any other previous work on in-ear EEG.
OpenEarable ExG features a NINA-B306-00B Bluetooth-capable module, featuring a single core ARM Cortex-M4 microcontroller. OpenEarable ExG also incorporates a 9-axis Inertial Measurement unit consisting of an accelerometer, a gyroscope, and a magnetometer. 
A microSD card can be used to save data locally, but in our work it was directly streamed over BLE. For the biopotential measuring chain, an analog-to-digital converter combined with instrumentation amplifiers is integrated. 

\begin{figure}[h]
    \centering
    \includegraphics[width=0.5\textwidth]{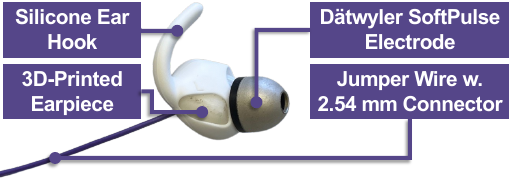}
    \caption{Earpiece (left ear) used for the sleep study, consisting of a 3D-printed earpiece with a silicone ear hook and Dätwyler SoftPulse electrodes.}
    \label{fig:earpiece}
\end{figure}

The electrode earpieces used, depicted in \autoref{fig:earpiece}, are an adaptation of the open-source design provided by \citet{lepold_openearable_2024}; they are flatter and therefore more comfortable to wear. Additionally, silicone ear hooks can be attached to the earpieces, to secure them better in the ear during sleep.
The single size, generic electrode tips are manufactured by Dätwyler\footnote{https://datwyler.com/company/innovation/softpulse}. 

\subsection{Reference device}
Our reference device is the Apple Watch Ultra. Its sleep classification algorithm relies on 3-axis accelerometer data to label 30-second epochs as \textit{Awake}, \textit{Core} (N1/N2), \textit{Deep} (N3), or \textit{REM} \cite{estimating_2023}, following the AASM sleep staging convention.
Apple validated the algorithm using over 1,400 nights of data from polysomnography (PSG) and at-home EEG recordings scored by trained professionals. It achieved high sensitivity for sleep detection (97.8\%) and moderate specificity (76\%), with an average Cohen’s kappa of 0.63. Most classification errors occurred between physiologically similar stages, such as core and deep sleep. Clinical settings showed slightly lower kappas (0.55–0.57), but sensitivity remained high.
Independent studies confirm these findings. \citet{robbins_accuracy_2024} reported that the Apple Watch Series 8 had high sensitivity (97\%) but overestimated light sleep by 45 minutes and underestimated deep sleep by 43 minutes ($\kappa$ = 0.60). \citet{miller_validation_2022} found that the Series 6 detected 97\% of sleep epochs but only 26\% of wake epochs, with an overall $\kappa$ of 0.30. \citet{roomkham_sleep_2019} observed close agreement with a validated actigraphy device (r = 0.85), with minor overestimation of total sleep time.

While not clinically certified, we have specifically chosen the Apple Watch over PSG, as it offers a good balance between usability and accuracy in an unsupervised, at-home setup. For this study, it served as a reference to assess the sleep detection capabilities of the OpenEarable ExG system.

\subsection{Recording Setup}
A total of eleven participants (4 females, 7 males) were recruited for the sleep experiment. Participant ages ranged from 21 to 59 years, with a median age of 24. All participants provided informed consent; data were anonymized, and the study was conducted in accordance with our university’s institutional ethics guidelines and the Declaration of Helsinki. Among the eleven participants, eight recorded one night of sleep, two recorded two nights, and one recorded three nights, resulting in a total of 15 nights’ worth of recordings. This variation was due to the participants’ differing levels of comfort sleeping with the device attached.

\begin{figure}[t]
    \centering
    \includegraphics[width=0.5\textwidth]{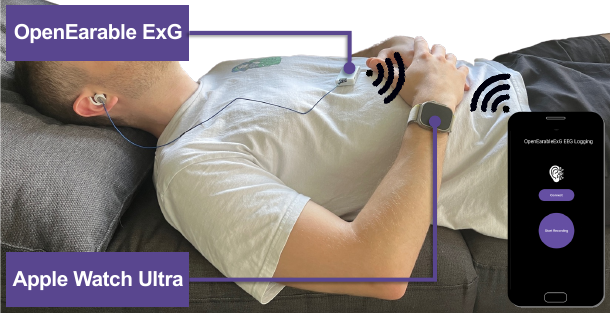}
    \caption{Overview of the sleep experiment. A participant is wearing the OpenEarable ExG with in-ear electrodes, resting on his chest, and an Apple Watch Ultra. OpenEarable ExG streams data to a corresponding app via Bluetooth.}
    \label{fig:Overview of the sleep experiment}
\end{figure}

Participants were instructed to wear the OpenEarable ExG during sleep, with both in-ear electrodes inserted and the Apple Watch worn on the wrist.
Biopotential data were sampled at approximately 250 Hz and streamed from the OpenEarable ExG via Bluetooth Low Energy (BLE) to a custom-developed smartphone application, where it was recorded locally.
In total, the recordings yielded approximately 87.5 hours of sleep data, corresponding to 12,004 epochs prior to further data processing.
The application also recorded the sleep staging data of the Apple Watch.
After receiving instructions, participants independently set up the experiment and used the smartphone application to start and stop the recordings.

\subsection{Data Processing and Feature Extraction}
The raw biopotential data is filtered with a fourth-order Butterworth bandpass filter (0.5–30 Hz) implemented in a stable second-order sections (SOS) format. The data is then aligned with the sleep stage labels obtained from the Apple Watch Ultra. Finally, the data is segmented into 30-second epochs, each assigned the dominant sleep stage, resulting in a filtered and labelled dataset suitable for further analysis.

Epochs were manually excluded if one or both earphones had dislodged during sleep or if the signal amplitudes exceeded $\pm$ 500 µV, to account for excessive movement artifacts.
This resulted in 7,099 remaining epochs, of the 12,004 recorded epochs. 
Only 2\% of the rejected epochs were due to movement artifacts, with the majority attributed to earphones falling out during sleep. Most movement-related artifacts occurred during wake periods.

A subset of features proposed by \citet{mikkelsen_accurate_2019} was implemented, with some modifications and additional features introduced to better suit our data:

\begin{itemize}
    \item \textbf{Time-Domain Features}: Standard deviation, variance, skewness, kurtosis, zero-crossing rate, and the 75\textsuperscript{th} percentile were computed to characterize the amplitude distribution and variability of the signal. Hjorth mobility and Hjorth complexity capture the signal’s frequency characteristics and complexity in the time domain.
    
    \item \textbf{Spectral features}: Relative power in classical EEG bands (delta: 0.5–4 Hz, theta: 4–8 Hz, alpha: 8–12~Hz, beta: 12–30~Hz) was computed using Welch’s method. Several power ratios were also derived:
    $[\delta/\theta, \theta/\alpha, \alpha/\beta$, as well as $(\theta~+~\delta)/(\alpha~+~\beta)]$.
    
    \item \textbf{Frequency-Domain Features}: Spectral entropy, spectral edge frequency, peak frequency, median frequency, and the difference between mean frequencies were included to summarize the spectral distribution.
    
    \item \textbf{Wavelet-Based Features}: Continuous wavelet transform (CWT) coefficients were computed in the delta, theta, alpha, and beta bands. For each band, the 75\textsuperscript{th} percentile of the absolute CWT coefficients was extracted.

    \item \textbf{Complexity Measure}: Lempel-Ziv complexity was used to estimate the signal's algorithmic complexity.
    
\end{itemize}

The importance of individual features was assessed using the built-in feature importance ranking of a random forest model, evaluated through a 10-fold cross-validation. To incorporate temporal context, a sliding window technique was applied: for each epoch, features from the two preceding and two succeeding epochs were concatenated with those of the center epoch, resulting in a feature vector that captures short-term temporal dependencies important for sleep stage transitions. The window size was set to five epochs (two before, one center, two after), and the final dataset consists of these extended feature vectors along with the label corresponding to the center epoch.

\subsection{Automatic Sleep Scoring}
We implemented a sleep classifier using two training and testing strategies: (i) 10-fold stratified cross-validation (10f-CV) and (ii) leave-one-participant-out cross-validation (LOPO-CV).
For both approaches, a random forest model was selected and implemented using the scikit-learn machine learning library.
Two classification strategies were developed: a multilabel classifier distinguishing between the sleep stages \textit{Awake}, \textit{REM}, \textit{Core}, and \textit{Deep}, and a binary classifier differentiating between \textit{Awake} and \textit{Asleep}, where \textit{Deep}, \textit{Core}, and \textit{REM} stages were merged into a single \textit{Asleep} label.
To address the under-representation of the \textit{Awake} class, SMOTE oversampling was applied using the implementation from the imbalanced-learn Python library. In the LOPO-CV setting, features were constructed using a sliding window approach to incorporate temporal context. To prevent data leakage, this windowing was not applied during 10-fold cross-validation.


\section{Results}

\begin{figure}[H]
    \centering
    \includegraphics[width=0.5\textwidth]{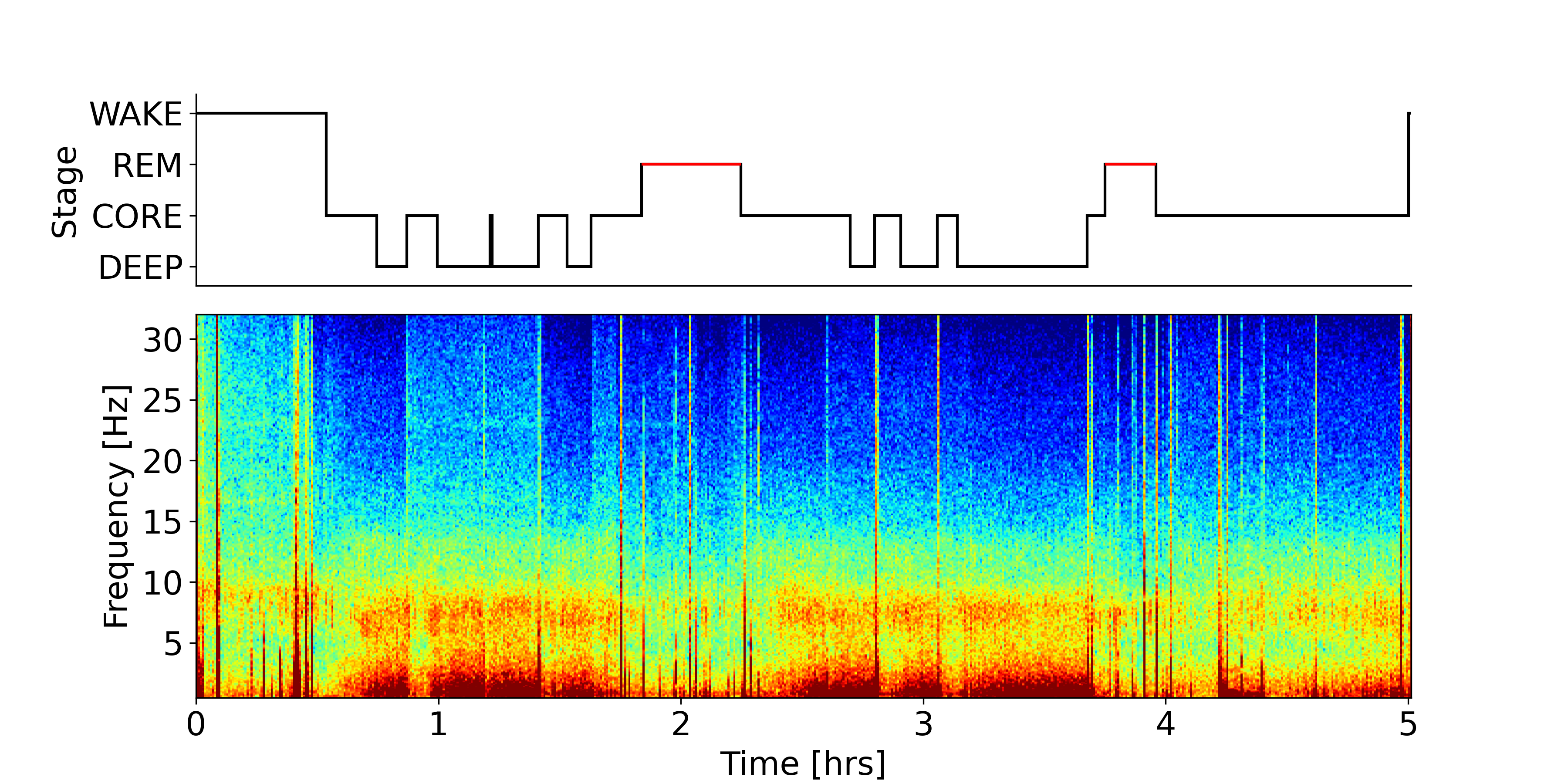}
    \caption{Spectrogram of ExG signal recorded with OpenEarable (bottom) with sleep stages derived from Apple Watch sleep scoring shown above (top).}
    \label{fig:spectrogram}
\end{figure}

\subsection{Sleep Phenomena}
We used the work of \citet{vallat_open-source_2021} to visualize and analyze the recorded signals. \autoref{fig:spectrogram} shows a spectrogram of the signal recorded using OpenEarable ExG (bottom), with sleep stages classified by the Apple Watch (top) based on its built-in sleep scoring. Distinct sleep-related patterns are visible upon visual inspection: delta wave (0.5–2 <Hz) activity increases throughout the core and deep sleep stages, with a more pronounced increase during deep sleep compared to core sleep. In contrast, the REM phase is marked by a clear reduction in delta activity, closely resembling a wake-like state.

\subsection{Classifier Performance}

\begin{figure*}[h]
\centering

\begin{subfigure}[t]{0.24\textwidth}
    \centering
    \includegraphics[width=1.1\linewidth]{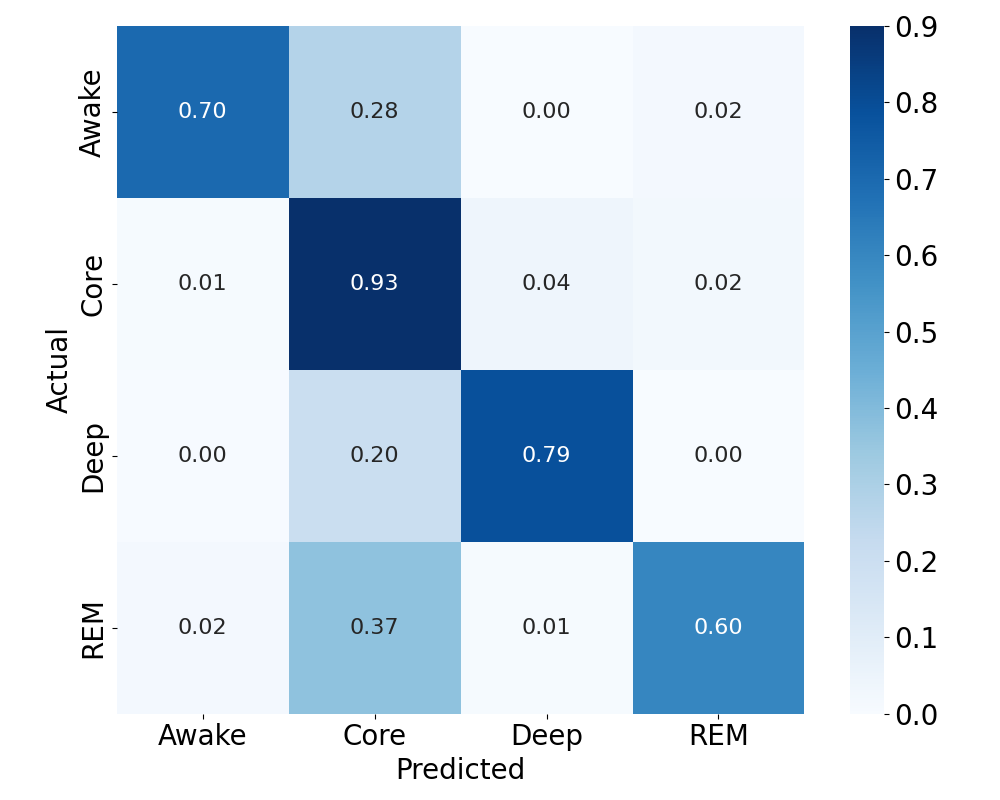}
    \caption{Multistage: 10-fold cross-validation.}
    \label{fig:multi_10cv}
\end{subfigure}
\hfill
\begin{subfigure}[t]{0.24\textwidth}
    \centering
    \includegraphics[width=1.1\linewidth]{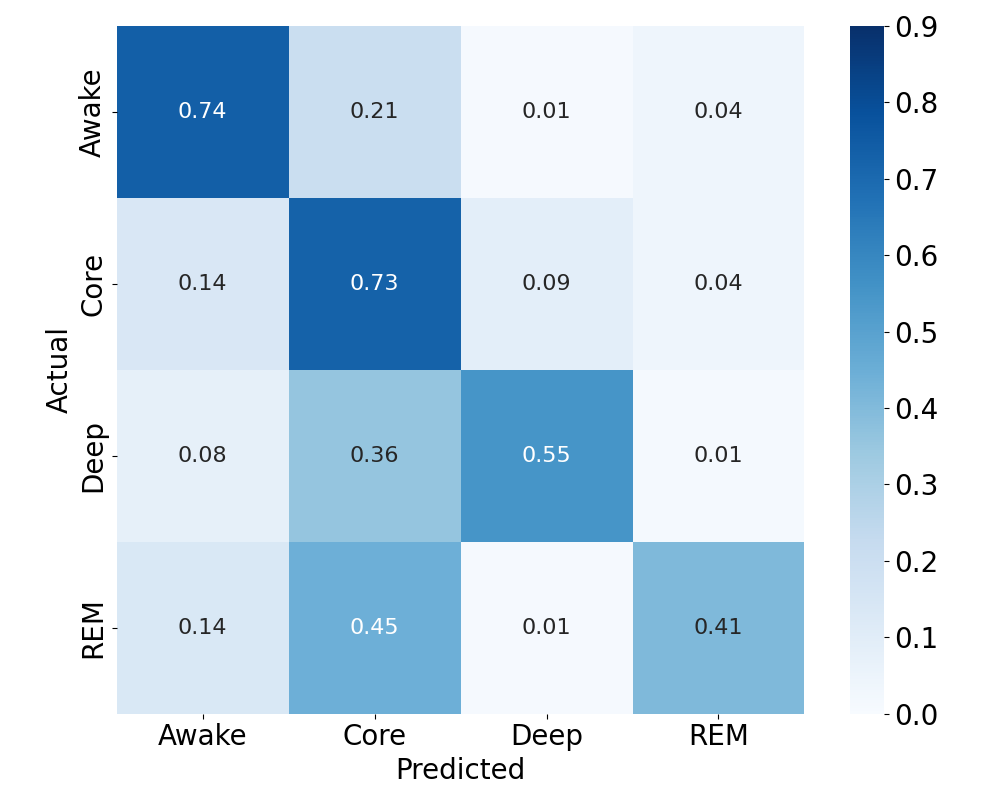}
    \caption{Multistage: leave-one-participant-out CV.}
    \label{fig:multi_lopo}
\end{subfigure}
\hfill
\begin{subfigure}[t]{0.24\textwidth}
    \centering
    \includegraphics[width=1.0\linewidth]{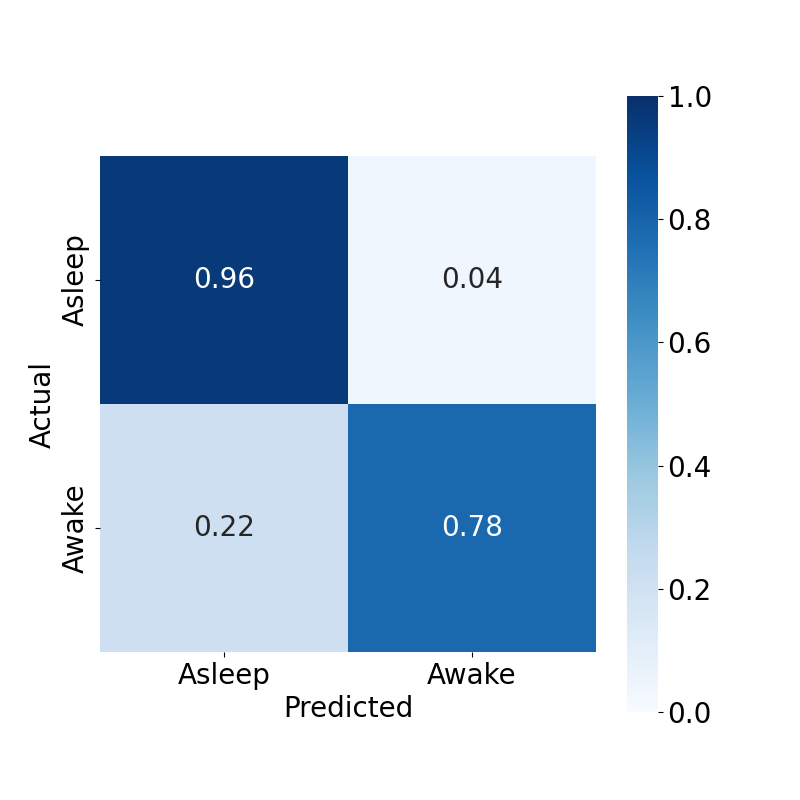}
    \caption{Binary: 10-fold cross-validation.}
    \label{fig:binary_10cv}
\end{subfigure}
\hfill
\begin{subfigure}[t]{0.24\textwidth}
    \centering
    \includegraphics[width=1.0\linewidth]{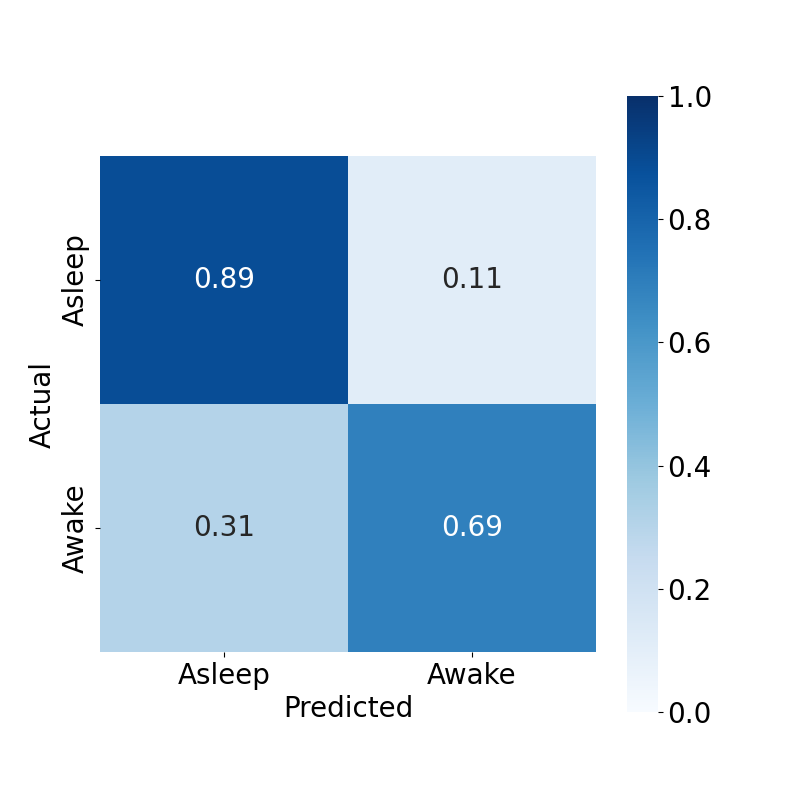}
    \caption{Binary: leave-one-participant-out CV.}
    \label{fig:binary_lopo}
\end{subfigure}

\vspace{1.5em}

\begin{subfigure}[t]{0.24\textwidth}
    \vspace{0pt}
    \centering
    \footnotesize
    \begin{tabular}{@{}lcc@{}}
    \toprule
    \textbf{Class} & \textbf{Sens.} & \textbf{Spec.} \\ \midrule
    Awake & 0.702 & 0.991 \\
    Core  & 0.929 & 0.723 \\
    Deep  & 0.790 & 0.969 \\
    REM   & 0.601 & 0.983 \\ \midrule
    \textbf{Overall} & \textbf{Acc: 0.840} & \textbf{Kappa: 0.710} \\
    \bottomrule
    \end{tabular}
    \caption{Multistage: 10-fold CV metrics.}
    \label{tab:multistage_metrics_10f}
\end{subfigure}
\hfill
\begin{subfigure}[t]{0.24\textwidth}
    \vspace{0pt}
    \centering
    \footnotesize
    \begin{tabular}{@{}lcc@{}}
    \toprule
    \textbf{Class} & \textbf{Sens.} & \textbf{Spec.} \\ \midrule
    Awake & 0.659 & 0.912 \\
    Core  & 0.709 & 0.674 \\
    Deep  & 0.600 & 0.887 \\
    REM   & 0.371 & 0.971 \\ \midrule
    \textbf{Overall} & \textbf{Acc: 0.651} & \textbf{Kappa: 0.443} \\
    \bottomrule
    \end{tabular}
    \caption{Multistage: leave-one-participant-out CV metrics.}
    \label{tab:multistage_metrics_lopo}
\end{subfigure}
\hfill
\begin{subfigure}[t]{0.24\textwidth}
    \vspace{0pt}
    \centering
    \footnotesize
    \begin{tabular}{@{}lcc@{}}
    \toprule
    \textbf{Class} & \textbf{Sens.} & \textbf{Spec.} \\ \midrule
    Awake  & 0.785 & 0.964 \\
    Asleep & 0.964 & 0.785 \\ 
    \\
    \\ \midrule
    \textbf{Overall} & \textbf{Acc: 0.948} & \textbf{Kappa: 0.701} \\
    \bottomrule
    \end{tabular}
    \caption{Binary: 10-fold CV metrics.}
    \label{tab:binary_metrics_10fold}
\end{subfigure}
\hfill
\begin{subfigure}[t]{0.24\textwidth}
    \vspace{0pt}
    \centering
    \footnotesize
    \begin{tabular}{@{}lcc@{}}
    \toprule
    \textbf{Class} & \textbf{Sens.} & \textbf{Spec.} \\ \midrule
    Awake  & 0.611 & 0.929 \\
    Asleep & 0.929 & 0.611 \\ 
    \\
    \\ \midrule
    \textbf{Overall} & \textbf{Acc: 0.905} & \textbf{Kappa: 0.515} \\
    \bottomrule
    \end{tabular}
    \caption{Binary: leave-one-participant-out CV metrics.}
    \label{tab:binary_metrics_lopo}
\end{subfigure}

\caption{Confusion matrices (top row) and corresponding performance metrics (bottom row) for sleep stage classification using multistage and binary models. Results are shown for both 10-fold cross-validation and leave-one-participant-out cross-validation (LOPO). The multistage model classifies four sleep stages (Awake, Core, Deep, REM), while the binary model distinguishes between Awake and Asleep states.}

\end{figure*}

First, we evaluate the performance of our sleep scoring model in classifying the recorded signal into the stages \textit{Awake}, \textit{Core}, \textit{Deep}, and \textit{REM}. \autoref{fig:multi_10cv} presents the confusion matrix (CF) for the 10-fold cross-validation (10f-CV) and \autoref{fig:multi_lopo} CF for the leave-one-participant-out cross-validation (LOPO-CV) approaches. The corresponding key performance metrics are summarized in \autoref{tab:multistage_metrics_10f} and \autoref{tab:multistage_metrics_lopo}, respectively.
Overall, the classification performance is lower for LOPO-CV compared to 10f-CV. In both approaches, the model struggles to accurately classify the wake stage, with frequent misclassification of \textit{Awake} as \textit{Core}, and vice versa. This issue is more pronounced in LOPO-CV. Additionally, distinguishing between \textit{Core} and \textit{Deep} stages becomes more challenging in the LOPO-CV approach. \textit{REM} is consistently the most difficult stage to classify in both validation approaches.

\begin{figure}[h!]
    \centering
    \begin{subfigure}[b]{0.5\textwidth}
        \centering
        \includegraphics[width=\textwidth]{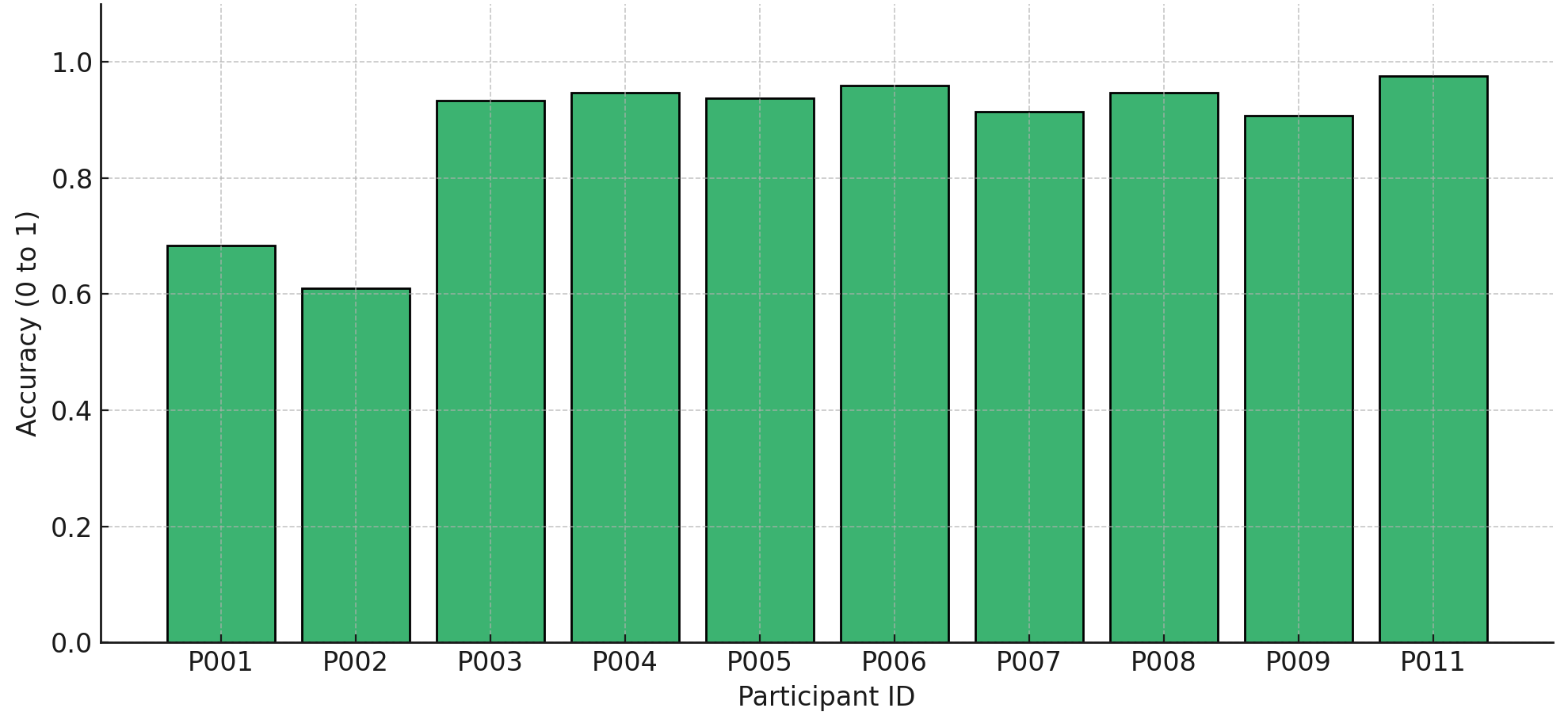}
        \caption{Accuracy per participant.}
        \label{fig:accuracy}
    \end{subfigure}
    \hfill
    \begin{subfigure}[b]{0.5\textwidth}
        \centering
        \includegraphics[width=\textwidth]{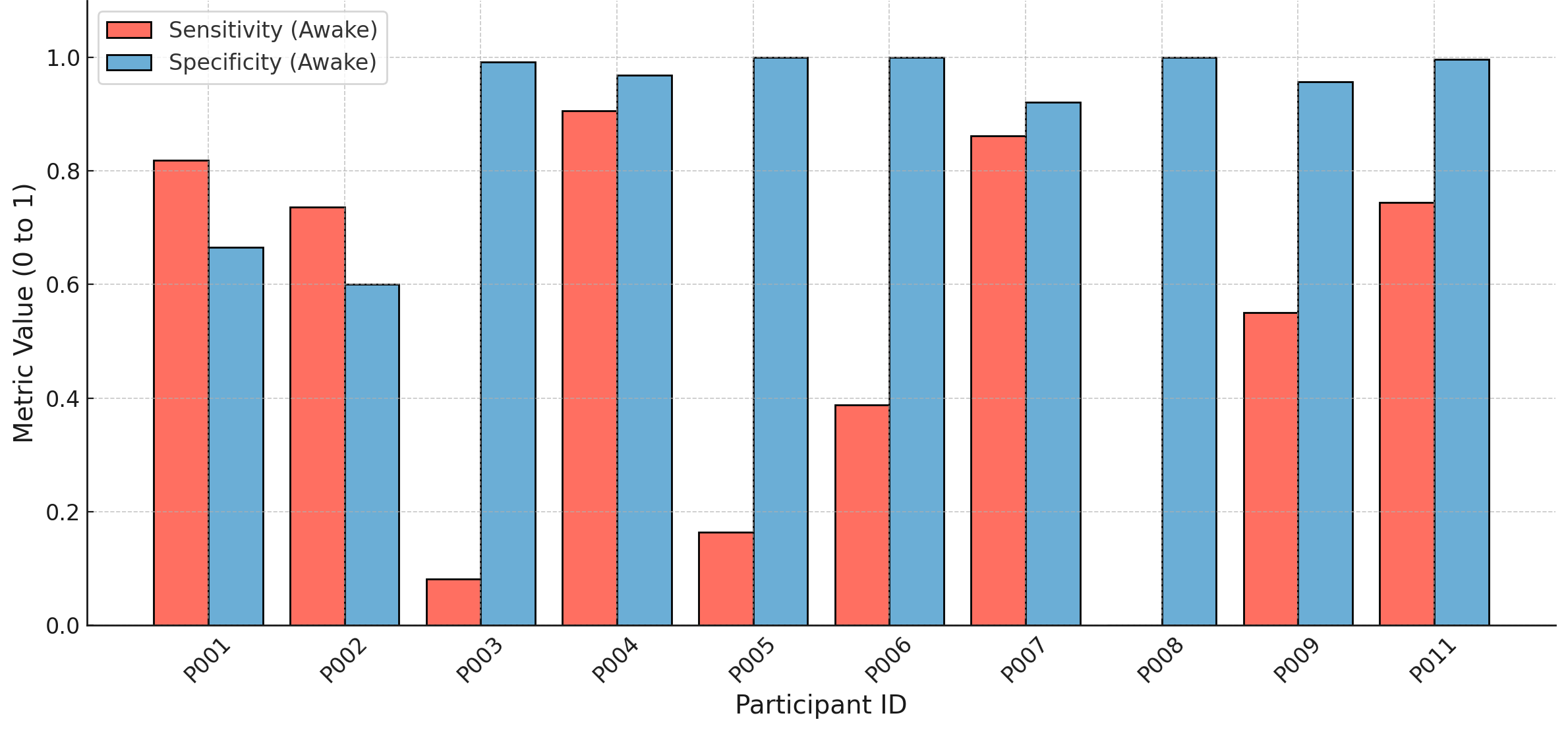}
        \caption{Sensitivity and specificity per participant.}
        \label{fig:sensitivity_specificity}
    \end{subfigure}
    \hfill
    \begin{subfigure}[b]{0.5\textwidth}
        \centering
        \includegraphics[width=\textwidth]{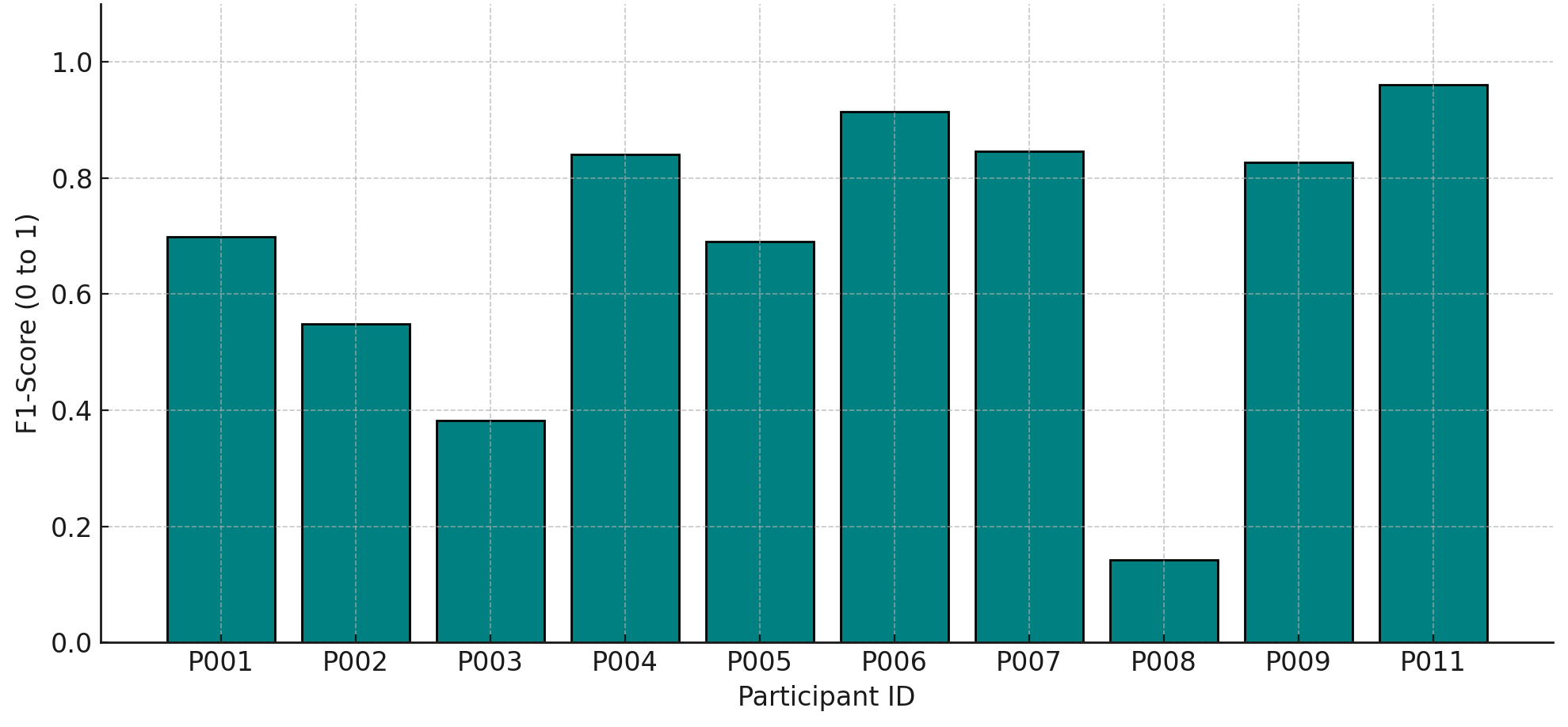}
    \caption{F1-Score per participant.}
    \label{fig:f1score_participant}
    \end{subfigure}

    \caption{Model performance metrics for binary classification (\textit{Awake} vs \textit{Asleep}) across all 11 participants, including accuracy, sensitivity \& specificity and F1-score.}
    \label{fig:combined_metrics}
\end{figure}


Therefore, we simplified the multistage classification into a binary task by combining the stages \textit{Core}, \textit{Deep}, and \textit{REM} into a single stage labeled \textit{Asleep} and retrained the model to determine whether the recorded signal indicated that the subject was awake or asleep.
\autoref{fig:binary_10cv} presents the confusion matrix (CF) for the 10-fold cross-validation (10f-CV) and \autoref{fig:binary_lopo} CF for the leave-one-participant-out cross-validation (LOPO-CV) approaches. The corresponding key performance metrics are summarized in \autoref{tab:binary_metrics_10fold} and \autoref{tab:binary_metrics_lopo}, respectively.
Similarly to multistage classification, the 10-fold cross-validation (10f-CV) approach outperformed leave-one-participant-out cross-validation (LOPO-CV). Although the difference between the two approaches in classifying the \textit{Asleep} class is marginal, the performance gap for the \textit{Awake} class is more substantial.
The corresponding confusion matrices are presented in \autoref{fig:binary_10cv} and \autoref{fig:binary_lopo}, with detailed performance metrics shown in \autoref{tab:binary_metrics_10fold} and \autoref{tab:binary_metrics_lopo}.
Again, the reason for this is the significant under-representation of the \textit{Awake} class.

Visualizing the classifier performance for individual participants provides insight into the overall lower performance of LOPO-CV compared to the 10-fold CV. Although the classifier performed perfectly for some participants, it did not correctly classify the majority of epochs for others. 
Figure \autoref{fig:combined_metrics} illustrates the per-participant accuracy, sensitivity/specificity, and F1-scores for binary classification, highlighting the considerable variability in classification performance.
This suggests that our dataset is too limited to adequately capture the variability in sleep phenotypes, leading to challenges in generalizing from the training set to certain recordings. This is expected, since sleep characteristics vary from person to person.

\begin{figure}[h]
    \centering
    \includegraphics[width=0.5\textwidth] {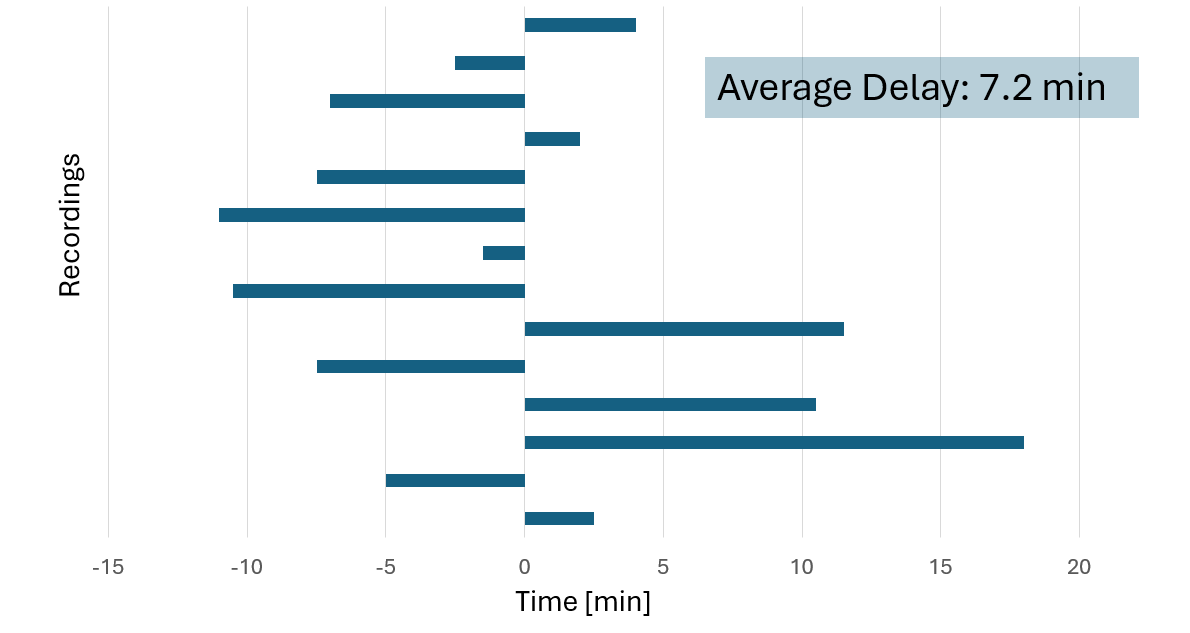}
    \caption{Difference in detected sleep onset times between the sleep classifier and Apple Watch across recordings. Each bar represents one recording, with negative values indicating earlier detection by the classifier and positive values indicating later detection. The delays ranged from -11 to +18 minutes, with a mean absolute difference of 7.2 minutes.}
    \label{fig:comparison_sleep_latency}
\end{figure}

To assess the practical performance of the sleep classifier, we compared its predicted sleep onset times with those recorded by the Apple Watch. Specifically, we examined the time points at which each device first identified the wearer as asleep. \autoref{fig:comparison_sleep_latency} compares the difference between the two devices for each recording. A negative time difference indicates that the classifier underestimated the duration of the wake stage, while a positive time difference signifies an overestimation. The observed delays ranged from -11 minutes to +18 minutes. The mean absolute delay, calculated across all recordings, was 7.2 minutes. 

\subsection{Participant Experience and Feedback}

During post-experiment interviews and a follow-up questionnaire, participants reported several issues related to wearing the earphones during sleep. Some noted that the earphones caused discomfort and pain in the ear (N=8), especially for side sleepers. Participants highlighted the need for a more ergonomic design to improve comfort, suggesting either better-fitting earpieces, or the ability to remove one earphone to accommodate different sleeping positions. While a few participants expressed interest in using the system for automatic media playback stopping or sleep tracking (N=4), most preferred alternative devices due to the discomfort of wearing in-ear headphones overnight. 

Additional use cases suggested by participants included automatically turning off lights, stopping media playback, setting alarms, detecting drowsiness while driving, and silencing phone notifications during sleep. Although two participants raised concerns about potential long-term ear health effects, data privacy was generally not viewed as an issue. Overall, comfort was the dominant concern (N=6) and the main barrier to broader acceptance of ear-EEG devices for sleep applications.

\section{Conclusion and Future Work}

This study investigated the feasibility of using a single-channel, in-ear ExG device for automatic sleep staging in naturalistic, at-home settings. Our results indicate that such a system can achieve reasonable accuracy for binary sleep-wake classification using a lightweight classification pipeline and open-source hardware. While these initial findings are promising, they should be interpreted with caution given the small sample size, limited staging resolution, and reliance on a consumer-grade device for ground truth.

The broader aim of enabling unobtrusive and accessible sleep monitoring is compelling, but significant challenges remain. In particular, the modest multiclass staging performance, reports of user discomfort, and the absence of clinical validation highlight the need for further methodological refinement. Future work should prioritize improved ergonomic design, more rigorous validation against clinical standards, and comparative analysis with existing wearable systems.

Overall, this study offers a preliminary step toward exploring the potential of in-ear ExG for sleep monitoring but does not yet establish the approach as a robust alternative to current methods. Continued investigation is required to assess its viability in both research and applied contexts.
In sum, our findings contribute to a growing body of research that seeks to democratize sleep health technologies and move toward scalable, real-world sleep monitoring beyond the clinic.

\bibliographystyle{ACM-Reference-Format}
\balance
\bibliography{references}


\end{document}